\pdfoutput=1

\documentclass[11pt]{article}

\usepackage{acl}

\usepackage{times}
\usepackage{latexsym}

\usepackage[T1]{fontenc}

\usepackage[utf8]{inputenc}

\usepackage{microtype}

\newcommand{\ie}{\textit{i.e.\ }}
\newcommand{\eg}{\textit{e.g.\ }}
\newcommand{\etc}{\textit{etc.\ }}

\usepackage{amssymb}
\usepackage{mathtools}
\usepackage{booktabs}
\usepackage{multirow}
\usepackage{adjustbox}
\usepackage{subcaption}
\usepackage{graphicx}
\usepackage{appendix}
\usepackage{caption}
\usepackage{todonotes}

\title{Guiding Visual Question Generation}

\author{Nihir Vedd, Zixu Wang, Marek Rei, Yishu Miao, \and Lucia Specia \\
Imperial College London \\
\{n.vedd19, zixu.wang, marek.rei, y.miao20, l.specia\}@imperial.ac.uk \\
\url{https://github.com/nihirv/guiding-vqg}}

\begin{document}
\maketitle\begin{abstract}
In traditional Visual Question Generation (VQG), most images have multiple concepts (\eg objects and categories) for which a question could be generated, but models are trained to mimic an arbitrary choice of concept as given in their training data. This makes training difficult and also poses issues for evaluation -- multiple valid questions exist for most images but only one or a few are captured by the human references.
We present Guiding Visual Question Generation - a variant of VQG which conditions the question generator on categorical information based on expectations on the type of question and the objects it should explore. We propose two variant families: (i) an explicitly guided model that enables an actor (human or automated) to select which objects and categories to generate a question for; and (ii) 2 types of implicitly guided models that learn which objects and categories to condition on, based on discrete variables. 
The proposed models are evaluated on an answer-category augmented VQA dataset and our quantitative results show a substantial improvement over the current state of the art (over 9 BLEU-4 increase). Human evaluation validates that guidance helps the generation of questions that are grammatically coherent and relevant to the given image and objects.


\end{abstract}

\section{Introduction}
\label{sec:intro}

In the last few years, the AI research community has witnessed a surge in multimodal tasks such as Visual Question Answering (VQA) \citep{VQA, Anderson_2018_CVPR}, Multimodal Machine Translation \citep{specia-etal-2016-shared, elliott-etal-2017-findings, barrault-etal-2018-findings, caglayan-etal-2019-probing}, and Image Captioning (IC) \citep{Vinyals2015ShowAT, Karpathy_2015_CVPR, pmlr-v37-xuc15}. Visual Question Generation (VQG) \citep{Zhang2016AutomaticQuestions, Krishna2019InformationGeneration, li2018iqan}, a multimodal task which aims to generate a question given an image, remains relatively under-researched despite the popularity of its textual counterpart. Throughout the sparse literature in this domain, different approaches have augmented and/or incorporated extra information as input. For example, \citet{Pan2019RecentGeneration} emphasised that providing the ground truth answer to a target question is beneficial in generating a non-generic question. 
\citet{Krishna2019InformationGeneration} pointed out that requiring an answer to generate questions violates a realistic scenario. 
Instead, they proposed a latent variable model using answer categories to help generate the corresponding questions. 
Recently, \citet{Scialom2020WhatGeneration} incorporated a pre-trained language model with object features and image captions for question generation.



In this work, we explore VQG from the perspective of `guiding' a question generator. 
Guiding has shown success in image captioning (\citet{Zheng2018IntentionObjects} and \citet{Ng2020UnderstandingDomains}), and in this VQG work we introduce the notion of `guiding' as conditioning a generator on inputs that match specific chosen properties from the target. 
We use the answer category and objects/concepts based on an image and target question as inputs to our decoder. We propose our {\bf explicit guiding} approach to achieve this goal. We additionally investigate an {\bf implicit guiding} approach which attempts to remove the dependency on an external actor (see more below).

The explicit variant (Section \ref{sec:explicit}) is modelled around the notion that an actor can select a subset of detected objects in an image for conditioning the generative process. Depending on the application, this selection could be done by a human, and algorithm or chosen randomly.
For example, imagine either a open-conversation chat-bot or a language learning app. In the chat-bot case, a human may show the bot a picture of something. The bot may use randomly sampled concepts from the image (e.g. an object-detected tree) to ask a human a question upon.
In the language learning case, the human may wish to select 
certain concepts they want the generated question to reflect. For example, they might select a subset of animal-related objects from the whole set of detected objects in order to generate questions for teaching the animal-related vocabulary in a language learning setting. Alongside the objects, the actor may also provide, or randomly sample, an answer category to the question generator.

The implicit variant (Section \ref{sec:implicit}), on the other hand, is motivated by removing the dependency on the aforementioned actor.
We provide two methodologies for our proposed implicit variant. The first uses a Gumbel-Softmax \citep{Jang2016CategoricalGumbel-Softmax} to provide a discrete sample of object labels that can be used for generating a question. The second method employs a model with two discrete latent variables that learn an internally-predicted category and a set of objects relevant for the generated question, optimised with cross-entropy and variational inference \citep{Kingma2014Auto-encodingBayes, pmlr-v48-miao16}.

Human evaluation shows that our models can generate realistic and relevant questions, with our explicit model \textit{almost} fooling humans when asked to determine which, out of two questions, is the generated question. Our experiments and results are presented in Section \ref{sec:results}.

To summarise, our {\bf main contributions} are: 1) The first work to explore guiding using object labels in Visual Question Generation;
2) A novel generative Transformer-based set-to-sequence approach for Visual Question Generation; 3) The first work to explore discrete variable models in Visual Question Generation; and 4) A substantial increase in quantitative metrics - our explicit model improves the current state of the art setups by over 9 BLEU-4 and 110 CIDEr.


\begin{figure*}[!ht]
\centering
\begin{subfigure}[b]{1\textwidth}
   \includegraphics[width=1\linewidth]{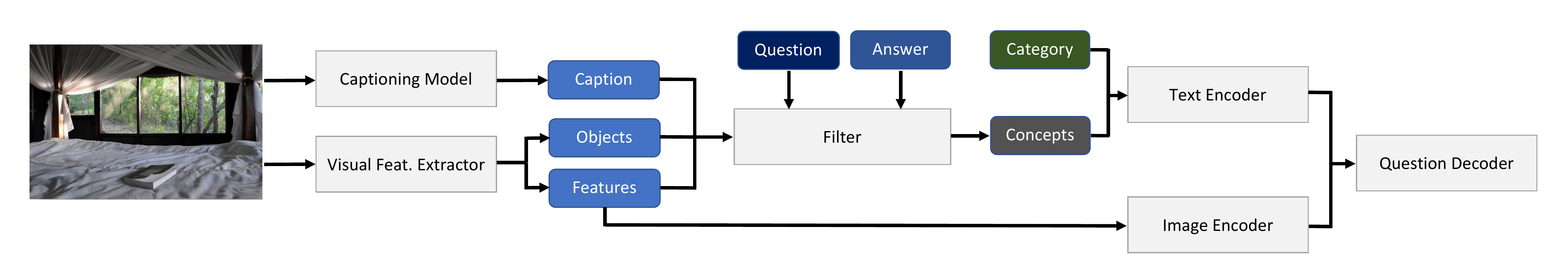}
   \caption{Architecture of our explicit model. Given an image, first an object detection model is used to extract object labels and object features; a captioning model is used to generate relevant captions. Questions and answers are concatenated to filter the conceptual information from generated objects and captions. Next the filtered concepts are combined with the category as the input to the text encoder; the extracted object features are fed into an image encoder. Finally the outputs from the text encoder and the image encoder are fused into the decoder for question generation. }
   \label{fig:Ng1} 
\end{subfigure}
\begin{subfigure}[b]{1\textwidth}
   \includegraphics[width=1\linewidth]{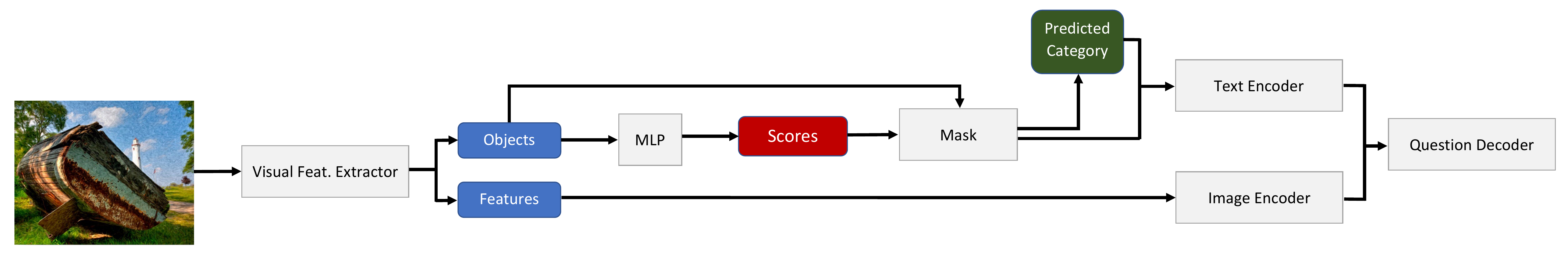}
   \caption{Architecture of our implicit model. Similar to the explicit model, first an object detection model is used to extract object labels and object features. Object labels are sent to a non-linear MLP after which a Gumbel-Softmax is applied to obtain the discrete vector `Scores'. The Scores are then used to mask the object labels and predict a category. The masked object labels and predicted category are then sent to the text encoder. The outputs are fused with the image encoder outputs and sent to the decoder.}
   \label{fig:non-variational-implicit} 
\end{subfigure}
\begin{subfigure}[b]{1\textwidth}
  \includegraphics[width=1\linewidth]{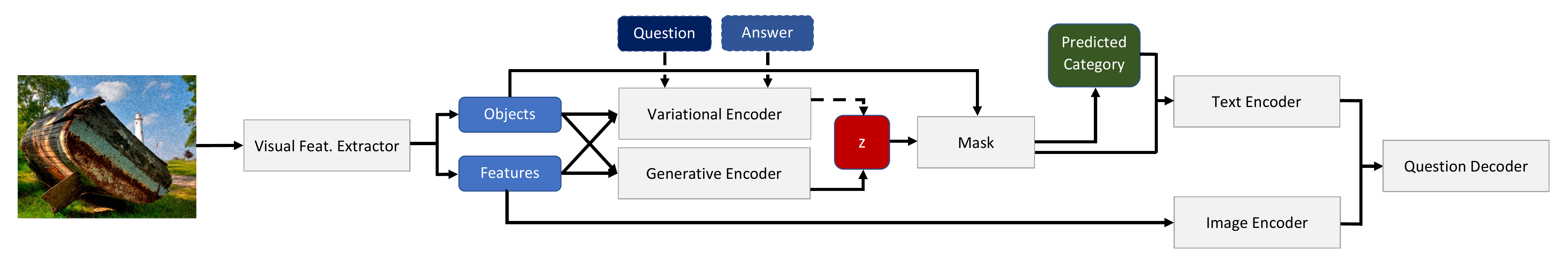}
  \caption{Architecture of our variational implicit model. After the object detection model extracts the object labels and object features, they are sent to the variational and generative encoders. The variational encoder is used at train time only, and also receives the question and answer pair. Depending whether we're training or in inference, we obtain a discrete vector $z$ from the respective distribution. $z$ is then used to mask the object labels. This variant then follows the same methodology as its non-variational counterpart. For this sub-figure only, the dashed lines indicate training.}
  \label{fig:Ng2}
\end{subfigure}
\caption{Architecture of the explicit model (a) and implicit model (b)}
\label{fig:architecture}
\end{figure*}

\section{Related Work}
\subsection{Visual Question Generation}\label{sec:related-vqg}
\citet{Zhang2016AutomaticQuestions} introduced the first paper in the field of VQG, employing an RNN based encoder-decoder framework alongside model-generated captions to generate questions. Since then, only a handful of papers have investigated VQG. \citet{Fan2018ABi-discriminators} demonstrated the successful use of a GAN in VQG systems, allowing for non-deterministic and diverse outputs. \citet{Jain2017Creativity:Autoencoders} proposed a model using a VAE instead of a GAN, however their improved results require the use of a target answer during inference. To overcome this unrealistic requirement, \citet{Krishna2019InformationGeneration} augmented the VQA \cite{VQA} dataset with answer categories, and proposed a model which doesn't require an answer during inference. Because their architecture uses information from the target as input (i.e. an answer category), their work falls under our definition of guided generation. More recently, \citet{Scialom2020WhatGeneration} investigate the cross modal performance of pre-trained language models by fine-tuning a BERT \cite{Devlin2018} model on model-based object features and ground-truth image captions. Other work, such as \citet{Patro2018MultimodalGeneration}, \citet{Patro2020DeepGeneration} and \citet{Uppal2020C3VQG:Generation}, 
either do not include BLEU scores higher than BLEU-1, which is not very informative,
or  address variants of the VQG task. In the latter case the models fail to beat previous SoTA on BLEU-4 for standard VQG. Recently and \cite{xu-2021:radian-gcn-for-vqg} and \cite{xie-2021:multiple-object-aware-vqg} achieve SoTA in VQG using graph convolutional networks. However, both works follow an unrealistic setup by conditioning their model on raw answers during training and inference - a dependency we attempt to remove. 

\subsection{Discrete (Latent) Variable Models}
Discrete variable models are ideal for tasks which require controllable generation \citep{Hu2017TowardText} or `hard' indexing of a vector \citep{Graves2016HybridMemory}. Existing literature provide several methods to achieve discretization. NLP GAN literature (such as SeqGAN \citep{Yu2016SeqGAN:Gradient} and MaskGAN \citep{Fedus2018MaskGAN:The______}) commonly use REINFORCE \citep{Williams1992SimpleLearning} to overcome differentiability issues with discrete outputs. Other discretization methodologies can be found in Variational Auto Encoder (VAE) literature \cite{Kingma2014Auto-encodingBayes}. Some older methodologies are NVIL \citep{Mnih2014NeuralNetworks} and VIMCO \citep{Mnih2016VariationalObjectives}. However, VAE literature also introduced Concrete \citep{Maddison2016TheVariables}, Gumbel-Softmax \citep{Jang2016CategoricalGumbel-Softmax} and Vector Quantization \citep{Oord2017NeuralLearning} as discretization strategies (technically speaking, Concrete and Gumbel-Softmax are strongly peaked continuous distributions).

In this work, we use a Gumbel-Softmax approach to sample a distribution over objects. At inference time, given a set of object tokens, learning this `hard' distribution allows the model to internally sample a subset of objects that produce the most informative question. Our variational model additionally learns a generative and variational distribution that allow the model to implicitly learn which objects are relevant to a question and answer pair whilst incorporating non-determinism for diverse outputs.

\section{Methodology}\label{sec:methodology}
We introduce the shared concepts of our explicit and implicit model variants, before diving into the variant-specific methodologies (Section \ref{sec:explicit} \& \ref{sec:implicit}). 

For both variants, we keep the VQG problem grounded to a realistic scenario. That is, during inference, we can only provide the model with an image, and data that can either be generated by a model (\eg object features or image captions) and/or trivially provided by an actor (\ie answer category and a selected subset of the detected objects). 
However, during training, we are able to use any available information, such as images, captions, objects, answer categories, answers and target questions, employing latent variable models to minimise divergences between feature representations of data accessible at train time but not inference time. This framework is inspired by \citet{Krishna2019InformationGeneration}. In Appendix \ref{app:tti}, we discuss the differences of input during training, testing and inference.

Formally, the VQG problem is as follows: Given an image $\tilde{i} \in \tilde{I}$, where $\tilde{I}$ denotes a set of  images, decode a question $q$. In the {\bf guided} variant, for each $\tilde{i}$, we also have access to textual utterances, such as ground truth answer categories and answers. The utterances could also be extracted by an automated model, such as image captions \cite{Li2020Oscar:Tasks}, or object labels 
and features \cite{Anderson_2018_CVPR}. In our work, answer categories take on 1 out of 16 categorical variables to indicate the type of question asked. For example, ``\textit{how many people are in this picture?}'' would have a category of ``\textit{count}'' (see  \citet{Krishna2019InformationGeneration} for more details). 

\textbf{Text Encoder.} For encoding the text, we use BERT \citep{Devlin2018} as a pre-trained language model (PLM). 
Thus, for a tokenised textual input $\tilde{S}$ of length $T$, 
we can extract a $d$-dimensional representation for $\tilde{s_t} \in \tilde{S}$: $X = \texttt{PLM}(\tilde{S}) \in \mathbb{R}^{T \times d}$

\textbf{Image Encoder.} Given an image $\tilde{i}$, we can extract object features, $f \in \mathbb{R}^{k_o \times 2048}$, and their respective normalized bounding boxes, $b \in \mathbb{R}^{k_o \times 4}$, with the 4 dimensions referring to horizontal and vertical positions of the feature bounding box. Following the seminal methodology of \citet{Anderson_2018_CVPR}, $k_o$ is usually 36. Subsequent to obtaining these features, we encode the image using a Transformer \cite{Vaswani2017}, replacing the default position embeddings with the spatial embeddings extracted from the bounding box features \cite{fairseqImageTransformer,Cornia2019Meshed-MemoryCaptioning}. 
Specifically, given $f, b$ from image $\tilde{i}$: $i = \texttt{Transformer}(f, b) \in \mathbb{R}^{k_o \times d}$


\textbf{Text Decoder.} We employ a pretrained Transformer decoder for our task \citep{wolf-etal-2020-transformers}. Following standard sequence-to-sequence causal decoding practices, our decoder receives some encoder outputs, and auto-regressively samples the next token, for use in the next decoding timestep. Our encoder outputs are the concatenation ($;$ operator) of our textual and vision modality representation: $X = [S; i] \in \mathbb{R}^{(T+k_o) \times d}$, and 
our decoder takes on the form: $\hat{q} = \texttt{Decoder}(X)$,
where $\hat{q}$ is the predicted question. 

In this work, we primarily focus on a set-to-sequence problem as opposed to a sequence-to-sequence problem. That is, our textual input is not a natural language sequence, rather an unordered set comprising of tokens from the answer category, the object labels, and the caption. How this set is obtained is discussed in following section. Due to the set input format, we disable positional encoding on the PLM encoder (Text Encoder in Figure \ref{fig:architecture}). 


\subsection{Explicit Guiding}
\label{sec:explicit}
As mentioned in Section \ref{sec:intro}, the explicit variant requires some actor in the loop. 
Thus, in a real world setting, this model will run in two steps. Firstly, we run object detection (OD) and image captioning (IC) over an image and return relevant guiding information to the actor. The actor may then select or randomly sample a subset of objects which are sent to the decoder to start its generation process. If the actor opts for a random sample strategy, no human is needed during the inference process (see Appendix \ref{app:tti} for examples).

To enable this setup, we create paired data based on the guided notion. At a high level, our approach creates this data in three steps: 1) obtain object labels; 2) obtain concepts via IC
Formally,
%
\begin{equation}\label{eq:oc-candidates}
\begin{split}
        &\text{objects} = \texttt{OD}(i) \in \mathbb{R}^{k_o} \\
        &\text{cap} = \texttt{CaptionModel}(i) \in \mathbb{R}^{T_{cap}} \\
        &\text{cap} = \texttt{rmStopWords}(\text{caption}) \in \mathbb{R}^{<T_{cap}} \\
        &\text{candidate\_concepts} =  \texttt{set}(\text{objects}; \text{cap}) \in \mathbb{R}^{T_{cc}}
\end{split}
\end{equation}

Here, \texttt{OD} stands for an object detector model, \texttt{rmStopWords} is a function which removes the stop words from a list, and \texttt{set} is a function which creates a set from the concatenation (the ; operator) of the detected objects and obtained captions. \textit{cap} stands for caption. The set is of size $T_{cc} < k_o + T_{cap}$. Using this obtained \textit{candidate\_concepts} set, we run our filtration process.

Once the set of candidate concepts has been constructed, we filter them to only retain concepts relevant to the target QA pair. 
After removing stop words and applying the \texttt{set} function to the words in the QA pair, we use Sentence-BERT \cite{Reimers2019Sentence-BERT:BERT-Networks} to obtain embeddings for the candidate QA pair and \textit{candidate\_concepts} (Eq \ref{eq:oc-candidates}). 
We subsequently compute a cosine similarity matrix between the two embedding matrices, and then select the top $k$ most similar concepts. 
The chosen $k$ concepts, $\tilde{S}$, are always a \emph{strict subset} of the candidate concepts that are retrieved using automated image captioning or object detection.
This process emulates the selection of objects an actor would select in an inference setting when given a choice of possible concepts, and creates paired data for the guided VQG task. We now concatenate an answer category to $\tilde{S}$: $S = \texttt{PLM}([\tilde{S}; \text{category}]) \in \mathbb{R}^{T \times d}$.





With text encoding $S$, we run the model, optimizing the negative log likelihood between the predicted question and the ground truth. Note that the concatenation in the decoder below is along the sequence axis (resulting in a tensor $\in \mathbb{R}^{T+k_o \times d}$).
\begin{equation}
    \begin{aligned}
        &\hat{q} = \texttt{Decoder}([S; i]) \\ 
        &\mathcal{L} = \texttt{CrossEntropy}(\hat{q}, q)
    \end{aligned}
\end{equation}

\subsection{Implicit Guiding}
\label{sec:implicit}
We now introduce our experiments for the implicit variant for VQG. This variant differs from its explicit counterpart as it aims to generate questions using only images as the input, while internally learning to predict the relevant category and objects. Mathematically, the explicit variant models $\hat{q} = p(w_t | i, \tilde{S}, \textit{category}, w_0, ..., w_{t-1}; \theta)$ where $\tilde{S}$ and \textit{category} are obtained as described in Section \ref{sec:explicit}. During inference, the implicit variant instead attempts to model $\hat{q} = p(w_t | i, \tilde{e}_{obj}, e_{cat}, w_0, ..., w_{t-1}; \theta)$ where $\tilde{e}_{obj}, e_{cat}$ are \textbf{not} explicitly fed in to the model. Rather, they are determined internally as defined in Equation \ref{eq:implict-text-encoder-and-category-prediction}.

Given an image, we apply the same object detection model as in the explicit variants to extract object labels, which are then encoded using an \texttt{embed} layer. 
Formally,
\begin{equation}\label{eq:embed-object-labels}
    \begin{aligned}
        &\text{objects} = \texttt{OD}(i) \in \mathbb{R}^{k_o} \\
        &e_{obj} = \texttt{embed}(\text{objects}) \in \mathbb{R}^{k_o \times d} 
    \end{aligned}
\end{equation}

Since we would like the implicit model to learn relevant objects for an image internally, we project each object in $e_{obj}$ to a real-valued score:

\begin{equation}
    scores = \texttt{MLP}(e_{obj}) \in \mathbb{R}^{k_o}
\end{equation}

Subsequently, we apply a hard Gumbel-Softmax \citep{45822} to obtain predictions over selected objects. Because Gumbel-Softmax samples from a log-log-uniform distribution, stochasticity is now present in our sampled objects. To sample $k$ objects, we tile/repeat \textit{scores} $k$ times before inputting it into the Gumbel-Softmax. $\tilde{z}$, our $k$-hot sampled objects vector, is then used to mask object embeddings for use in decoding:

\begin{equation}
    \begin{aligned}
        &\tilde{z} = \texttt{gumbel-softmax}(\text{scores}, k) \in \mathbb{R}^{k_o} \\
        &\tilde{e}_{obj} = \tilde{z} * e_{obj} \in \mathbb{R}^{k_o \times d}
    \end{aligned}
\end{equation}

Where $*$ denotes element-wise multiplication. Categories can also be a strong guiding factor and instead of making it an explicit input, we build a classifier to predict possible categories. In this variant, $\tilde{e}_{obj}$ is used as an input to both our text encoder, and the MLP responsible for the category prediction:
\begin{equation}\label{eq:implict-text-encoder-and-category-prediction}
    \begin{aligned}
        &S = \texttt{PLM}(\tilde{e}_{obj}) \in \mathbb{R}^{k_o \times d} \\
        &p(\hat{cat}|\tilde{e}_{obj}) = \texttt{softmax}(\texttt{MLP}(\tilde{e}_{obj})) \in \mathbb{R}^{k_{cat}}
    \end{aligned}
\end{equation}

Using the one-hot representation of the predicted category (i.e. $e_{cat} = \texttt{one-hot}(p(\hat{cat}|\tilde{e}_{obj}))$, we can concatenate our image, PLM representation of objects, and predicted category to feed into the decoder: $\hat{q} = \texttt{Decoder}([i; S; e_{cat}]) \in \mathbb{R}^{T_{\hat{q}}}$. However, during training, we teacher force against the `gold' set of objects, $\tilde{S}$ (obtained using \textit{candidate\_concepts} in Equation \ref{eq:oc-candidates}). Training and optimization thus follow:

\begin{equation}\label{eq:implicit-loss}
    \begin{aligned}
        \hat{q} = &\texttt{Decoder}([i; \tilde{S}; e_{cat}]) \in \mathbb{R}^{T_{\hat{q}}} \\
        \mathcal{L} = &\texttt{CrossEntropy}(\hat{q}, q) + \\
        &\texttt{CrossEntropy}(p(\hat{cat}|\tilde{e}_{obj}), cat) + \\
        &\texttt{StartEnd}(\tilde{e}_{obj}, \tilde{S})
    \end{aligned}
\end{equation}

\noindent where StartEnd is a BERT QA-head style loss \citep{Devlin2018} that uses binary cross entropy for each $k$ in $\tilde{e}_{obj}$.

\textbf{Variational Implicit.} Hypothesising that ground-truth QA pairs might provide information useful to selecting objects, we additionally attempt to extend our model to incorporate QA pairs to learn a latent variational distribution over the objects. However, since QA pairs can only be used during training to learn a variational distribution, we introduce another generative distribution that is only conditioned on the images and extracted objects. We borrow the idea from latent variable models to minimise Kullback-Leibler (KL) divergence between the variational distribution and generative distribution, where the variational distribution is used during training and the generative distribution is used in inference.

Continuing from Equation \ref{eq:embed-object-labels}, we build two matrices, $M_{gen}$ and $M_{var}$. The former is a concatenation of the image features and object embeddings, and the latter the concatenation between the encoded QA pair and $M_{gen}$. Depending on whether we're in a training or inference regime, the CLS token of the relevant matrix is used to sample a mask, $\tilde{z}$, which is subsequently applied on the aforementioned object embeddings:
\begin{equation*}
    \begin{aligned}
        &M_{gen} = \texttt{encode}([e_{obj}; i]) \in \mathbb{R}^{2k_o \times d} \\
        &e_{qa} = \texttt{embed}(\text{Q;A}) \in \mathbb{R}^{T_{qa} \times d} \\
        &M_{var} = \texttt{encode}([e_{qa}; M_{gen}]) \in \mathbb{R}^{(2k_o + {T^{qa}) \times d}} \\
        &q_{\phi}(z|M_{gen}, M_{var}) = \texttt{MLP}(M^{CLS}_{gen}; M^{CLS}_{var}) \in \mathbb{R}^{k_o} \\
        &p_{\theta}(z|M_{gen}) = \texttt{MLP}(M_{gen}) \in \mathbb{R}^{k_o} \\
        &\tilde{z} = \texttt{gumbel-softmax} (z, k) \in \mathbb{R}^{k_o} \\
        &\tilde{e}_{obj} = \tilde{z} * e_{obj} \in \mathbb{R}^{k_o \times d}
    \end{aligned}
\end{equation*}

\noindent where $q_{\phi}(z|M_{gen}, M_{var})$ is the variational distribution, $p_{\theta}(z|M_{gen})$ is the generative distribution, and MLP denotes a multilayer perceptron for learning the alignment between objects and QA pairs. $\texttt{encode}$ is an attention-based function such as BERT \citep{Devlin2018}. From here, our methodology follows on from Equation \ref{eq:implict-text-encoder-and-category-prediction}. However, our loss now attempts to minimise the ELBO:
\begin{equation*}
    \begin{aligned}
        \mathcal{L} &= \mathbb{E}[\log p_{\theta}(\hat{q}|z, \hat{cat})] \\
        &- D_{\text{KL}}[q_{\phi}(z|M^{CLS}_{gen},M^{CLS}_{var}) || p_{\theta}(z|M^{CLS}_{gen})] \\
        &+ \log p(\hat{cat}|M_{var})
    \end{aligned}
\end{equation*}

\section{Experiments}
\subsection{Datasets}
We use the VQA v2.0 dataset\footnote{\url{https://visualqa.org/}}~\citep{VQA} (CC-BY 4.0), a large dataset consisting of all relevant information for the VQG task.
We follow the official VQA partition, with \ie 443.8K questions from 82.8K images for training, and 214.4K questions from 40.5K images for validation. 
Following \citet{Krishna2019InformationGeneration}, we report the performance on validation set as the annotated categories and answers for the VQA test set are not available.

We use answer categories from the annotations of \citet{Krishna2019InformationGeneration}. 
The top 500 answers in the VQA v2.0 dataset are annotated with a label from the set of 15 possible categories, which covers up the 82\% of the VQA v2.0 dataset; the other answers are treated as an additional category. 
These annotated answer categories include objects (\eg “mountain”, “flower”), attributes (\eg “cold”, “old”), color, counting, \etc 

We report BLEU \citep{papineni-etal-2002-bleu}, ROUGE \citep{lin-2004-rouge}, CIDEr \citep{Vedantam2015CIDErCI}, METEOR \citep{lavie-agarwal-2007-meteor}, and MSJ \citep{montahaei2019jointly} as evaluation metrics. The MSJ metric accounts for both the diversity of generated outputs, and the n-gram overlap with the ground truth utterances.

\subsection{Comparative Approaches}
We compare our models with four recently proposed VQG models \textit{Information Maximising VQG} (\textbf{IMVQG}; supervised with image and answer category) \citep{Krishna2019InformationGeneration}, \textit{What BERT Sees} (\textbf{WBS}; supervised with image and image caption) \citep{Scialom2020WhatGeneration}, \textit{Deep Bayesian Network} (\textbf{DBN}; supervised with image, scenes, image captions and tags/concepts) \citep{Patro2020DeepGeneration}, and \textit{Category Consistent Cyclic VQG} (\textbf{C3VQG}; supervised with image and answer category) \citep{Uppal2020C3VQG:Generation}.
We follow \textbf{IMVQG}'s evaluation setup because they hold the current SoTA in VQG for realistic inference regimes. We omit \cite{xu-2021:radian-gcn-for-vqg} and \cite{xie-2021:multiple-object-aware-vqg} from our table of results because these models follow an unrealistic inference regime, requiring an explicit answer during training and inference. Our baseline is an image-only model, without other guiding information or latent variables.

\subsection{Implementation Details}
In Section \ref{sec:methodology} we described the shared aspects of our model variants. The reported scores in Section \ref{sec:results} use the same hyperparameters and model initialisation. A table of hyperparameters and training details can be found in Appendix \ref{app:hparams}. BERT Base \cite{Devlin2018} serves as our PLM encoder and following \citet{wolf-etal-2020-transformers, Scialom2020WhatGeneration}, we use a pre-trained BERT model for decoding too. Though typically not used for decoding, by concatenating the encoder inputs with a \texttt{[MASK]} token and feeding this to the decoder model, we are able to obtain an output (\eg $\hat{q_1}$). This decoded output is concatenated with the original input sequence, and once again fed to the decoder to sample the next token. Thus, we use the BERT model as a decoder in an auto-regressive fashion. 

To encode the images based on the Faster-RCNN object features  \cite{Ren2015FasterNetworks, Anderson_2018_CVPR}, we use a standard Transformer \cite{Vaswani2017} encoder. Empirically, we find $k=2$ to be the best number of sampled objects.




\section{Results}
\label{sec:results}

\begin{figure*}
\centering
   \includegraphics[width=1\linewidth]{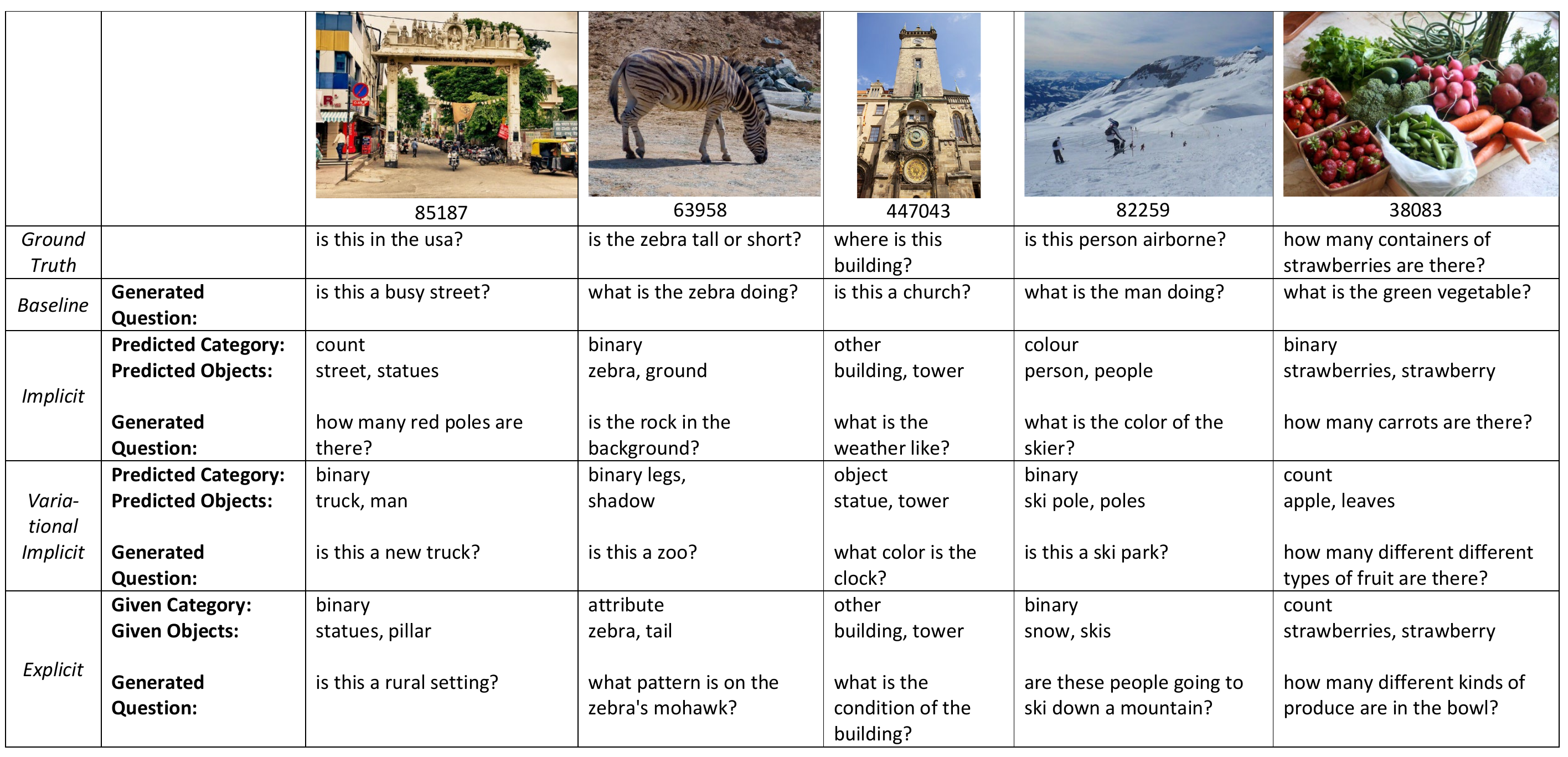}
\caption{Qualitative Examples. The ground truth is the target question for the baseline, implicit and explicit. The examples of explicit variant uses \texttt{image-guided} whereas the implicit is using the non-variational \texttt{image-pred}.}
\label{fig:qualitative}
\end{figure*}

We present quantitative results in Table \ref{tab:result-single} and qualitative results in Figure \ref{fig:qualitative}. We evaluate the explicit, implicit and variational implicit models in a single-reference setup, as the chosen input concepts are meant to guide the model output towards one particular target reference.


\begin{table*}[t]
\begin{center}
\resizebox{0.78\linewidth}{!}{
\begin{tabular}{@{}l|llllccllllc@{}}
\toprule
                         \multirow{2}{*}{Model}                                & \multicolumn{4}{c}{BLEU}                                                                      & \multicolumn{1}{c}{CIDEr} & \multicolumn{1}{c}{METEOR} & \multicolumn{1}{c}{ROUGE} & \multicolumn{3}{c}{MSJ}
                         \\ \cmidrule{2-12}
                                                        & \multicolumn{1}{c}{1} & \multicolumn{1}{c}{2} & \multicolumn{1}{c}{3} & \multicolumn{1}{c}{4} & \multicolumn{1}{c}{}      & \multicolumn{1}{c}{}       & \multicolumn{1}{c}{}      & \multicolumn{1}{c}{3} & \multicolumn{1}{c}{4} & \multicolumn{1}{c}{5}
                                            & \multicolumn{1}{c}{}
                                                        \\ \midrule
                                                        \multicolumn{11}{c}{\texttt{Comparative}} \\
 IMVQG (z-path)$^{\dagger}$          & \textbf{50.1}         & 32.3                  & 24.6                  & 16.3                  & 94.3                      & 20.6                       & 39.6                         & 47.2                    &    38.0                   &  31.5 \\
                         IMVQG (t-path)                  & 47.4                  & 29.0                  & 19.9                  & 14.5                  & 86.0                      & 18.4                       &  38.4                         &  53.8                     &  44.1                     &  37.2 \\
                             WBS$^{\ddagger}$ & 42.1                  & 22.4                  & 14.1                  & 9.2                   & 60.2                      & 14.9                       & 29.1                      &  63.2                     &  55.7                    &  49.7 \\
                              DBN & 40.7                  & -                  & -                  & -                   & -                      & 22.6                       & -                      & -                     & -                     & -     \\
                              C3VQG & 41.9                  & 22.1                  & 15.0                  & 10.0                   & 46.9                      & 13.6                       & 42.3                      & -                     & -                     & -\\
                             image-only                        & 25.9                  & 15.9                  & 9.8                  & 5.9                  & 41.4                      & 13.5                       & 27.8                      & 52.2                  & 42.8                  & 36.0\\ \midrule \midrule
\multicolumn{11}{c}{\texttt{Explicit}}   \\
image-category                    & 40.8                  & 29.9                  & 22.5                  & 17.3                  & 131                       & 20.8                       & 43.0                      & 64.2                  & 55.5                  & 48.8 \\
                              image-objects                     & 34.7                  & 25.0                    & 19.1                  & 15.0                    & 130                       & 19.4                       & 36.9                      & 67.4                  & 59.2                  & 52.7    \\
                              image-guided                     & 46.3                    & \textbf{36.4}         & \textbf{29.5}         & \textbf{24.4}         & \textbf{214}              & \textbf{25.2}                & \textbf{49.0}             & \textbf{71.3}         & \textbf{63.6}         & \textbf{57.3}  \\ 
                              image-guided-random                     & 23.6                    & 12.1         & 5.75         & 2.39         & 17.6              & 10.8                & 24.2             & 62.3         & 52.6         & 45.0\\ \midrule
\multicolumn{11}{c}{\texttt{Implicit}} \\   
image-category                    & 28.4                  & 17.5                  & 11.3                  & 8.5                   & 42.8                      & 13.5                         & 30.7                      & 51.8                  & 42.9                  & 36.4 \\
                             image-guided                    & 33.8                  & 24.0                  & 18.3                  & 14.2                  & 123                      & 19.1                       & 35.9                      & 66.7                  & 58.9                    & 52.5 \\
                             image-guided-pred                    & 25.3                  & 14.9                  & 9.1                  & 6.3                  & 27.3                      & 11.6                       & 27.3                      & 52.0                 & 44.0                    & 38.1 \\
                             image-guided-random                    & 21.3                  & 11.4                  & 6.3                  & 3.6                  & 23.1                      & 10.7                       & 22.2                      & 61.7                 & 52.8                    & 45.9 \\\midrule
\multicolumn{11}{c}{\texttt{Variational Implicit}} \\   
                             image-guided                    & 33.9                  & 23.5                  & 16.8                  & 12.6                  & 113                      & 18.8                       & 35.6                      & 64.2                  & 56.3                    & 49.8 \\
                             image-guided-pred                    & 22.6                  & 12.5                  & 6.9                  & 4.1                  & 24.3                      & 11.2                       & 23.0                      & 58.6                 & 49.3                    & 42.4 \\   
                             image-guided-random                    & 19.8                  & 10.7                  & 5.9                  & 3.3                  & 19.6                      & 10.0                       & 21.3                      & 58.8                 & 50                    & 43.4 \\\bottomrule
\end{tabular}
}
\end{center}
\caption{Single reference evaluation results. ``*-guided'' refers to the combination of category and objects. In the explicit variant only, objects refers to the subset of detected objects and caption keywords, filtered on the target QA pair. $^{\dagger}$ indicates an unrealistic inference regime, using answers as input for question generation. $^{\ddagger}$ WBS scores are from single reference evaluation based on the VQA1.0 pre-trained ``Im. + Cap.'' model provided by the authors. 
}
\label{tab:result-single}
\end{table*}

\subsection{Quantitative Results}
Starting with the explicit variant, as seen in Table \ref{tab:result-single}, 
we note that our image-only baseline model achieves a BLEU-4 score of 5.95. We test our model with different combinations of text features to identify which textual input is most influential to the reported metrics. We notice that the contribution of the category is the most important text input with respect to improving the score of the model, raising the BLEU-4 score by more than 11 points (image-category) over the aforementioned baseline. However, whilst the BLEU-4 for the image-object variant is 2.3 points lower, it outperforms the image-category variant by 3.9 points on the diversity orientated metric MSJ-5 - indicating that the image-category variant creates more generic questions. As expected, the inclusion of both the category and objects (image-guided)
outperforms either of the previously mentioned models, achieving a new state-of-the-art result of 24.4 BLEU-4. This combination also creates the most diverse questions, with an MSJ-5 of 57.3.

We also test our hypothesis that guiding produces questions that are relevant to the fed in concepts. This is tested with `image-guided-random' variant. This variant is the same trained model as `image-guided', but uses $k=2$ random concepts from a respective image instead of using the ground truth question to generate concepts. Our results show that guiding is an extremely effective strategy to produce questions related to conceptual information, with a BLEU-4 score difference of over 20 points. We refer the reader to Section \ref{sec:human-eval} for human evaluation which again validates this hypothesis, and Section \ref{sec:explicit} for an explanation of why guiding is valid for evaluating VQG models.

We evaluate the implicit models as follows. The implicit image-category variant does not predict any objects internally. It uses all image features and object embeddings alongside the category supervision signal as described in Equation \ref{eq:implicit-loss}. The implicit image-guided models use the `gold' objects at inference (See Section \ref{sec:explicit}). If these variants fit the `gold' objects well, it indicates that their generative abilities are suitable for guiding/conditioning on predicted or random objects. The image-guided-pred variants are evaluated using internally predicted objects - and the model variant that would be used in a real inference setting. Finally, the image-guided-random variants are fed in random object labels at inference. 

For implicit guiding to be a valid methodology, we need to validate two criteria: 1) Successfully conditioning the decoder on guiding information; 2) Better than random accuracy of object prediction/selection. Note that intuitively, the implicit model is expected to perform worse than the explicit model in terms of the language generation metrics. This is because of the inherently large entropy of the relevant answer category and the objects given an image. However, if the learned distributions over the categories and objects can capture the relevant concepts of different images, they may benefit the question generation when compared with image-only.

According to Table \ref{tab:result-single}, 
by predicting just an answer category and no objects (image-category), the proposed implicit model beats the image-only baseline. The BLEU-4 score difference is less than 1 with the best performing WBS model \cite{Scialom2020WhatGeneration} -- which also generates questions without explicit guided information.

As mentioned above, we can evaluate the implicit model by either feeding the `gold' objects obtained as described in Section \ref{sec:explicit}, or by the internally predicted objects as described in Section \ref{sec:implicit}. These form the variants image-guided and image-guided-pred respectively. For both the implicit and variational implicit models, image-guided is expected to perform the best. Results validate this, showing a performance of 14.2 and 12.6 BLEU-4 respectively. Importantly, the relatively high scores of these guided models (compared to the comparative approaches) show that these models can successfully be conditioned on guiding information.

We also notice that for both types of implicit models, image-guided-pred outperforms image-guided-random. Specifically for the non-variational implicit, we see a higher BLEU-4 score difference of 2.7. Interestingly, despite this BLEU-4 difference being higher than its variational counterpart, there is a trade-off for the diversity-orientated MSJ metric. This indicates that although generated questions are discretely `closer' to the ground truth, similar phrasing is used between the generated questions. In fact, an acute case of this phenomena occurs for the image-category variant where the BLEU-4 variant is higher than image-guided-pred or image-guided-random. In this case, qualitative analysis shows us that the higher BLEU-4 score can be attributed to the generic nature of the generated question. Failure cases of automatic evaluation metrics in NLP is discussed further in \citep{Caglayan2020CuriousTale}.

To satisfy the `better than random accuracy of object prediction/selection' criteria previously outlined, we measure the overlap of the $k$ predicted objects vs $k$ `gold' object labels. These `gold' object labels are obtained similarly to the explicit variant (Section \ref{sec:explicit}), however the caption tokens are not fed to the filtering process. Random accuracy for selecting objects is 12.5\%. Our overlap accuracy on implicit image-pred is 18.7\% - outperforming random selection. Variational implicit image-pred failed to outperform random accuracy.

\subsection{Qualitative Results}
Qualitative results are shown in Figure \ref{fig:qualitative} and Appendix \ref{app:qualitative}. Figure \ref{fig:qualitative} depicts how outputs from different model variants compare to ground truth questions. Without any guiding information, the image-only variant is able to decode semantic information from the image, however this leads to generic questions. The implicit variant, for which we also report the predicted category and objects, mostly generates on-topic and relevant questions. Focusing on the explicit variant, we witness high-quality, interesting, and on-topic questions. 

Appendix \ref{app:qualitative} depicts how well our explicit image-guided variant handles a random selection of detected objects given the image. This experiment intends to gauge the robustness of the model to detected objects which may fall on the low tail of the human generating question/data distribution. 
To clarify, humans are likely to ask commonsense questions which generally focus on obvious objects in the image. By selecting objects at random for the question to be generated on, the model has to deal with object permutations not seen during training, and categories that are invalid for an image.

Analysing the outputs, when viable categories and objects that are expected to fall in a commonsense distribution are sampled, the model can generate high quality questions. Interestingly, we observe that when the sampled objects are not commonsense (\eg ``ears arms'' for the baby and bear picture), the model falls back to using the object features instead of the guiding information. This phenomenon is also witnessed when the sampled category does not make sense for the image (\eg category `animal' in image 531086). Despite the category mismatch, the model successfully uses the object information to decode a question.


\subsection{Human Evaluation}\label{sec:human-eval}
\begin{table}[]
\begin{adjustbox}{width=0.48\textwidth}
\begin{tabular}{@{}l|llll@{}}
             & Baseline  & Implicit & V-Implicit  & Explicit \\ \midrule
Experiment 1 & 34.3\% $\pm$ 0.1 & 47.1\% $\pm$ 0.12 & 36.7\% $\pm$ 0.08 & 44.9\% $\pm$ 0.08 \\
Experiment 2 & 95.9\% $\pm$ 0.03 & 76.6\% $\pm$ 0.16 & 89\% $\pm$ 0.09 & 93.5\% $\pm$ 0.06 \\
Experiment 3 & -         & -         & - & 77.6\% $\pm$ 0.09 \\
Experiment 4 & -         & -         & - & 74.1\%/40.0\% $\pm$ 0.07/0.18 \\
\end{tabular}
\end{adjustbox}
\caption{Human evaluation results (and standard dev.)}
\end{table}
We ask seven humans across four experiments to evaluate 
the generative capabilities of our models. \textit{Experiment 1} is a visual Turing test: given an image, a model generated question and a ground truth question, we ask a human to determine which question they believe is model generated. \textit{Experiment 2} attempts to discern the linguistic and grammatical capabilities of our model by asking a human to make a binary choice about whether the generated question seems natural. \textit{Experiment 3} shows a human an image alongside a model generated question (explicit variant). Then, we ask the human to make a choice about whether the generated question is relevant to the image (\ie could an annotator have feasibly asked this question during data collection). Finally, \textit{experiment 4} judges whether objects are relevant to a generated question. The experiment is set up with true-pairs and adversarial-pairs. True-pairs are samples where the shown objects are the ones used to generate the question. Adversarial-pairs show a different set of objects than those which generated the question. If more true-pairs are are marked correct (\ie if at least one of the objects is relevant to the generated question) than the adversarial-pairs, then our model successfully generates questions on guiding information. 

In \textit{experiment 1}, a model generating human-level questions should be expected to score 50\%, as a human would not be able to reliably distinguish them from the manually created questions. Our results show the explicit and non-variational implicit model outperforming the variational implicit and baseline variants, fooling the human around 45\% of the time. Whilst not yet at the ideal 50\%, the explicit approach provides a promising step towards beating the visual Turing Test. \textit{Experiment 2} evaluates the grammatical plausibility of the generated questions. In general, all models perform extremely well in this experiment, with the baseline variant generating grammatically correct sentences ~96\% of the time. This is expected, as the baseline typically falls back to decoding easy/generic questions. \textit{Experiment 3}, is evaluated on our best performing model (explicit image-guided). Here, ~78\% of the generated questions are marked as relevant/on-topic given an image. Finally, \textit{experiment 4}'s results show true-pairs marked as correct vs adversarial-pairs (incorrectly) marked as correct. Since the former is larger than the latter - 72\% vs 42\%, the model can successfully use guiding/object information to create on-topic questions.

\section{Conclusions}
We presented a guided approach to visual question generation (VQG), which allows for the generation of questions that focus on specific chosen aspects of the input image. We introduced three variants for this task, the explicit, implicit, and variational implicit. The former generates questions based on an explicit answer category and a set of concepts from the image. In contrast, the latter two discretely predict these concepts internally, receiving only the image as input. The explicit model achieves SoTA results when evaluated against comparable models. Qualitative evaluation and human-based experiments demonstrate that both variants produce realistic and grammatically valid questions.

\section*{Acknowledgments}
Lucia Specia, Zixu Wang and Yishu Miao received support from MultiMT project (H2020 ERC Starting Grant No. 678017) and the Air Force Office of Scientific Research (under award number FA8655-20-1-7006).

\bibliography{anthology,custom,references}
\bibliographystyle{acl_natbib}

\clearpage
\appendix
\section{Training, testing and inference}\label{app:tti}
Here, using an example, we clarify the inputs to our explicit model (Section \ref{sec:explicit}) in the training, testing and inference setups. \\

\noindent\textbf{Training}

\begin{itemize}
    \item Ground truth question: What is the labrador about to catch?
    \item Answer: Frisbee
    \item Category: Object
    \item Image: $i \in \mathbb{R}^{k_o \times d}$
    \item \{Caption\}: A man throwing a frisbee to a dog
    \item \{Objects\}: person, dog, frisbee, grass
\end{itemize}

N.B. \{Caption\} and \{Objects\} are both model generated, requiring only an image as input. These inputs are thus available at inference time.

Firstly, we create a set of \emph{candidate\_concepts} (see eq. \ref{eq:oc-candidates}) from the caption and objects: [person, dog, frisbee, grass, man, throwing] ($\in \mathbb{R}^6$). These words are individually embedded. Secondly, we concatenate and embed the set of question and answer tokens ($\in \mathbb{R}^7$).

Then, we construct a matrix which gives us cosine similarity scores for each \emph{candidate\_concepts} token to a QA token ($\in \mathbb{R}^{6 \times 7}$). We choose $k=2$ tokens from the \emph{candidate\_concepts} which are most similar to the words from the QA. Here, “dog” and “frisbee” are likely chosen. Our input to the model is then <$i$, “object”, “dog”, “frisbee”>.

Notice that it is possible for these words to be in the QA pair (e.g. “frisbee”). Importantly, these words have not been fed from the QA pair - they have been fed in from model-obtained concepts (\{Object\} and \{Caption\}). Philosophically similar, \citet{Krishna2019InformationGeneration} constructed inputs based on target information for use in training and benchmarking.

\noindent\textbf{Testing.}
Imagine a data labeler creating questions based on an image. They would look at the image, and decide on the concepts to create the question for. Our testing methodology follows this intuition using the strategy outlined above: the $k=2$ selected objects from \emph{candidate\_concepts} is a programmatic attempt for selecting concepts which \emph{could} generate the target question. Note that there can be many questions generated for a subset of concepts (e.g. `is the dog about to catch the frisbee?', `what is the flying object near the dog?' etc.). As outlined above, we are not taking concepts from the target. Rather we use information from the target to emulate the concepts an actor would think of to generate the target question. Because there can be different concepts questions are based on for one image (see  ground-truth questions in Appendix \ref{app:qualitative}), our strategy allows us to generate questions which might be similar to a singular target question. This leads to an evaluation which fairly uses information a human has access to to generate a question.

\noindent\textbf{Inference.}
However, in the real world, there is no `ground-truth' question. In this case, we simply feed image features, and actor selected concepts to our question generator model. The selection process of the actor may be random - in which case a human agent does not need to be involved in the question generation process. The $k \leq 2$ selected concepts here are a subset of \emph{candidate\_concepts}, which are fully generated from models.

\section{Hyperparameters and training details} \label{app:hparams}
\begin{table}[h!]
\begin{tabular}{ll}
\hline
Batch size              & 128  \\
Learning rate           & 1e-5 \\
Text model layers       & 12   \\
Text model dimension    & 768  \\
Image encoder layers    & 6    \\
Image encoder dimension & 768  \\
Image encoder heads     & 8    \\ \hline
\end{tabular}
\caption{Hyperparameters for our model variants.}
\label{tab:hyperparams}
\end{table}

Empirically, for both variants, we find $k=2$ to be the best number of sampled objects. All experiments are run with early stopping (patience 10; training iterations capped at 35000) on the BLEU-4 metric. 
Scores reported (in Section \ref{sec:results}) are from the highest performing checkpoint. We use the PyTorch library and train our model on a V100 GPU (1.5 hours per epoch).


\section{Impact of model size on results} \label{app:model-size}
\begin{table}[h]
\begin{adjustbox}{width=0.48\textwidth}
\begin{tabular}{@{}l|llllccc@{}}
\toprule
            \multirow{2}{*}{Model}      & \multicolumn{4}{c}{BLEU}                                                                      & \multicolumn{1}{c}{CIDEr} & \multicolumn{1}{c}{METEOR} & \multicolumn{1}{c}{ROUGE} \\ \cmidrule{2-8}
                                                              & \multicolumn{1}{c}{1} & \multicolumn{1}{c}{2} & \multicolumn{1}{c}{3} & \multicolumn{1}{c}{4} & \multicolumn{1}{c}{}      & \multicolumn{1}{c}{}       & \multicolumn{1}{c}{}      \\ \midrule
image-category                    & 38.6                  & 28.4                  & 21.4                  & 16.2                  & 118                      & 19.9                       & 40.1                      \\
image-guided                    & 44.5                  & 34.4         & 27.4         & 22.1         & 197             & 24.6              & 47             \\ \bottomrule
\end{tabular}
\end{adjustbox}
\caption{Truncated models single reference evaluation results.}
\label{tab:result-truncated}
\end{table}
Our models use the heavier Transformers than previous SoTA we compare to. For example, \cite{Krishna2019InformationGeneration} use ResNet and RNNs for their image encoder and question generator ($\sim$18M parameters). Our models have between 200-300M parameters. To validate that our results are not purely attributable to model size, we train a truncated version of image-category and image-guided (explicit only). We truncate our models by using only the first and last layers of our BERT based encoders and decoders ($\sim$36M parameters). Our closest model to theirs is the (truncated) explicit image-category, which achieves a BLEU-4 of 16.2 as seen in Table \ref{tab:result-truncated} - an improvement of 1.7 BLEU-4 over IMVQG's \textit{t-path}. Even if we attribute 100\% of this score improvement to the pre-trained nature of the BERT models we use, our methodology still introduces a 5.9 BLEU-4 increase over the image-category combination (truncated image-guided achieves a BLEU-4 of 22.1).

\section{More Qualitative Examples.} \label{app:qualitative}
Examples can be seen in Figure \ref{fig:qualitative} (next page). When examined, we see that the generated question accurately uses the guiding category when the category is valid for the given image. For example, 531086/1 has animal as the sampled category. Because no animal is present in the image, this category isn't valid for the image. The generated question then correctly relies on the object labels and visual modality to generate a valid question given the image. Similarly for 490505/2. 

There are some cases where a sampled object/concept is not valid given an image. For example, at least one of the objects in 22929/1, 41276/1, 531086/2, 281711/1, 490505/1 is not valid. In this case the model usually relies on the other available guiding information, prioritising the category information (e.g. 531086/2). In rare cases, the model has failure cases where some of the valid sampled objects may not be used in the generated question (e.g. 293705/2 and 490505/2).

The concept extractor utilises a pre-trained image captioning model and object detector model. This may lead to an accumulation of downstream errors, especially if the data fed into the pre-trained models are from a significantly different data generating distribution than those used to train the model. In this erroneous case, the model will likely fallback to rely on the image modality and category information to produce a generic question (e.g. 22929/1, 22929/2, 531085/1, 293705/2).

\begin{figure*}[h]
\centering
  \includegraphics[clip, trim=0.5cm 6cm 0.5cm 0.5cm, width=1\textwidth]{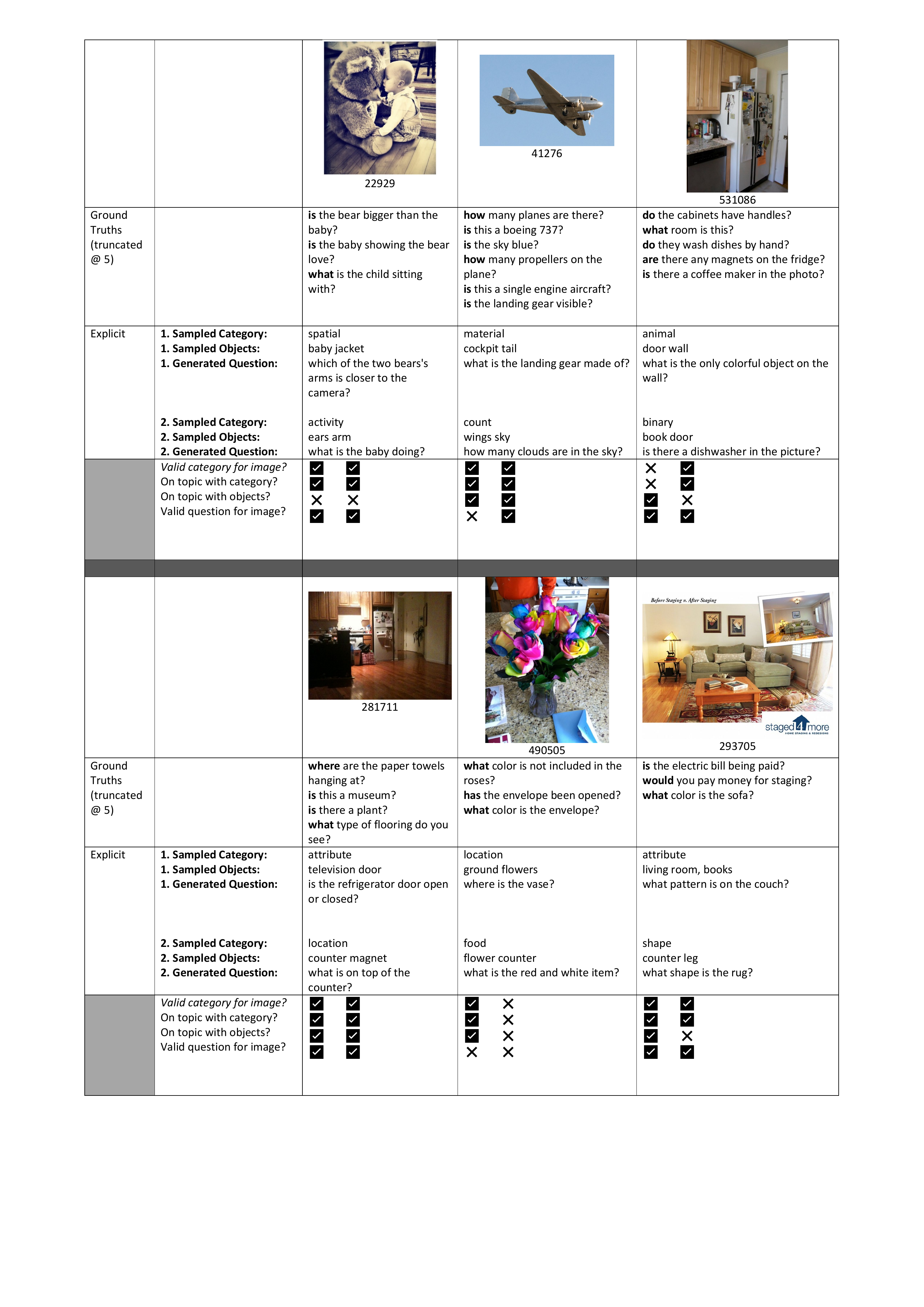}
\label{fig:qualitative2}
\caption{Qualitative outputs from explicit variant being fed random guiding information. Failure cases are also shown.}
\end{figure*}

\section{Responsible NLP Research}
\subsection{Limitations}
Our approach claims to achieve SoTA in Visual Question Generation. However, we are only able to train and test our model on one dataset because it is the only existing dataset which contains answer categories. It is possible that our work may be suitable for use in a zero-shot setting, but we have not evaluated or tested our model in this setup.

\subsection{Risks}
Our model could be used to generate novel questions for use in Visual Question Answering. This may have a knock-on effect which leads to training more VQA models, thus having a negative impact on the environment.

Our model could be used in downstream tasks such as language learning. There may be incorrectness in the generated questions which has a knock on effect to a user using this model (e.g. the user may gain a wrong understanding of a concept because of a question the model has generated) 

\end{document}